\documentclass[sigconf,authorversion,nonacm]{acmart}

\renewcommand\footnotetextcopyrightpermission[1]{} 
\usepackage{tabularx}
\usepackage{booktabs}
\usepackage{multirow}
\usepackage{tabularray} 
\usepackage{subcaption}
\AtBeginDocument{%
  \providecommand\BibTeX{{%
    \normalfont B\kern-0.5em{\scshape i\kern-0.25em b}\kern-0.8em\TeX}}}

\begin{document}
\pagestyle{plain}

\title[iSpLib: A Library for Accelerating Graph Neural Networks]{iSpLib: A Library for Accelerating Graph Neural Networks \\ using Auto-tuned Sparse Operations 
}


\author{Md Saidul Hoque Anik}
\affiliation{%
\department{Department of Intelligent Systems Engineering}
  \institution{Indiana University Bloomington}
  \city{Bloomington, Indiana}
  \country{USA}}
\email{mdshoque@iu.edu}

\author{Pranav Badhe}
\affiliation{%
\department{Department of Intelligent Systems Engineering}
  \institution{Indiana University Bloomington}
  \city{Bloomington, Indiana}
  \country{USA}}
\email{pbadhe@iu.edu}

\author{Rohit Gampa}
\affiliation{%
\department{Department of Intelligent Systems Engineering}
  \institution{Indiana University Bloomington}
  \city{Bloomington, Indiana}
  \country{USA}}
\email{rgampa@iu.edu}

\author{Ariful Azad}
\affiliation{%
\department{Department of Intelligent Systems Engineering}
  \institution{Indiana University Bloomington}
  \city{Bloomington, Indiana}
  \country{USA}}
\email{azad@iu.edu}

\renewcommand{\shortauthors}{Anik et al.}

\begin{abstract}
  Core computations in Graph Neural Network (GNN) training and inference are often mapped to sparse matrix operations such as sparse-dense matrix multiplication (SpMM).
  These sparse operations are harder to optimize by manual tuning because their performance depends significantly on the sparsity of input graphs, GNN models, and computing platforms.
  To address this challenge, we present iSpLib, a PyTorch-based C++ library equipped with auto-tuned sparse operations. 
  iSpLib expedites GNN training with a cache-enabled backpropagation that stores intermediate matrices in local caches. The library offers a user-friendly Python plug-in that allows users to take advantage of our optimized PyTorch operations out-of-the-box for any existing linear algebra-based PyTorch implementation of popular GNNs (Graph Convolution Network, GraphSAGE, Graph Inference Network, etc.) with only two lines of additional code. 
  We demonstrate that iSpLib obtains up to 27x overall training speedup compared to the equivalent PyTorch 2.1.0 and PyTorch Geometric 2.4.0 implementations on the CPU. Our library is publicly available at \href{https://github.com/HipGraph/iSpLib}{https://github.com/HipGraph/iSpLib}\footnote{\href{https://doi.org/10.5281/zenodo.10806511}{https://doi.org/10.5281/zenodo.10806511}}.
\end{abstract}









\maketitle  
\let\thefootnote\relax\footnotetext{Accepted as a short paper at the Web Conference 2024.}

\section{Introduction}
 Over the last decade, graph neural networks (GNNs) have demonstrated remarkable effectiveness in learning from graph-structured data.
This success has led to the development of several prominent libraries for GNNs, such as PyTorch Geometric (PyG)\cite{fey2019fast}, DGL\cite{wang2019deep}, and CogDL \cite{cen2021cogdl}, among others. Across these libraries, the fundamental computations for GNN forward and backward propagation primarily rely on two sparse operations: (a) sampled dense-dense matrix multiplication (SDDMM) and (b) sparse-dense matrix multiplication (SpMM). Consequently, the overall performance of these libraries is often influenced significantly by the efficiency of SDDMM and SpMM.
To enhance the speed of these sparse operations, we developed a general-purpose C++ library known as iSpLib. This library can be seamlessly integrated into the backend of PyTorch-based GNN training, facilitating accelerated GNN training and inference on various multi-core processors.

iSpLib takes advantage of the insight that high-level operations within GNNs, such as graph convolutions and message passing, 
can be mapped to sparse linear algebra.
Consequently, the library optimizes backend computations while preserving the familiar Python interface for users. iSpLib is designed with four key objectives:

{\bf(a) General-purpose:} iSpLib supports various GNN models, including GCN\cite{kipf2016semi}, GraphSAGE \cite{hamilton2017inductive}, and GIN \cite{xu2018powerful}, utilizing operations such as SpMM, SDDMM, and their combination known as FusedMM \cite{rahman2021fusedmm}. To accommodate a broad spectrum of GNNs, we break down the overall computation into smaller matrix and vector kernels. These micro kernels allow users to define their own operations.

{\bf(b) Fast and scalable GNN training and inference:} iSpLib significantly accelerates GNN training compared to existing sparse libraries like PyTorch-sparse \cite{rusty1s-no-date}. This speed is achieved through a harmonious combination of efficient parallelization, thread scheduling, loop unrolling, register blocking, and data caching during backpropagation.

{\bf(c) Performance portability:} iSpLib exhibits consistent performance across diverse multi-core processors without requiring users to manually optimize their code. Automated tuning based on different hardware features, such as SIMD intrinsics, vector lengths, and register sizes, enables iSpLib to generate optimized code for the target platform.

{\bf(d)  Easy-to-use library}: iSpLib seamlessly integrates into PyTorch-based libraries, such as PyG \cite{fey2019fast}, allowing users to divert computations to iSpLib with just one or two lines of code. This approach enables users to leverage efficient and auto-tuned kernels without explicit modifications to their existing code.

At present, iSpLib exclusively supports CPU-based tuning for Intel, AMD, and ARM processors. Our testing involved iSpLib's compatibility with PyG and various standalone GNN models, including the PyTorch-based GCN developed by the authors. We observed significant performance improvements when iSpLib was integrated with PyG, resulting in a GNN training speedup of up to $27\times$ for GCN, $12\times$ for GraphSAGE-sum, $8\times$ for GraphSAGE-mean, and $18\times$ for Graph Isomorphism Network (GIN). Importantly, these speed enhancements were achieved while maintaining the same level of accuracy attained without iSpLib and concealing all low-level optimizations from the users.

\section{Related Works}
Researchers used various approaches to expedite GNN training time. SparseTIR \cite{ye2023sparsetir} reports up to 7.45x speedup for sparse convolution using compilation abstraction. Lenadora et al. \cite{lenadora2023input} proposes a data-driven adaptive strategy and reports up to 1.99x speedup on graph convolutional networks. You et al. \cite{qiu2021optimizing} focus on choosing a suitable sparse matrix storage format to improve the GNN training performance and report up to 3x performance improvement in GNN running time. Wang et al. \cite{hu2020featgraph} proposes FeatGraph that accelerates GNN training by co-optimizing graph traversal and feature dimension computation and reports up to 32x speedup on CPU.

\section{Library Design}
\subsection{Overview}
\begin{figure}[h]
    
      \includegraphics[width=1.05\linewidth]{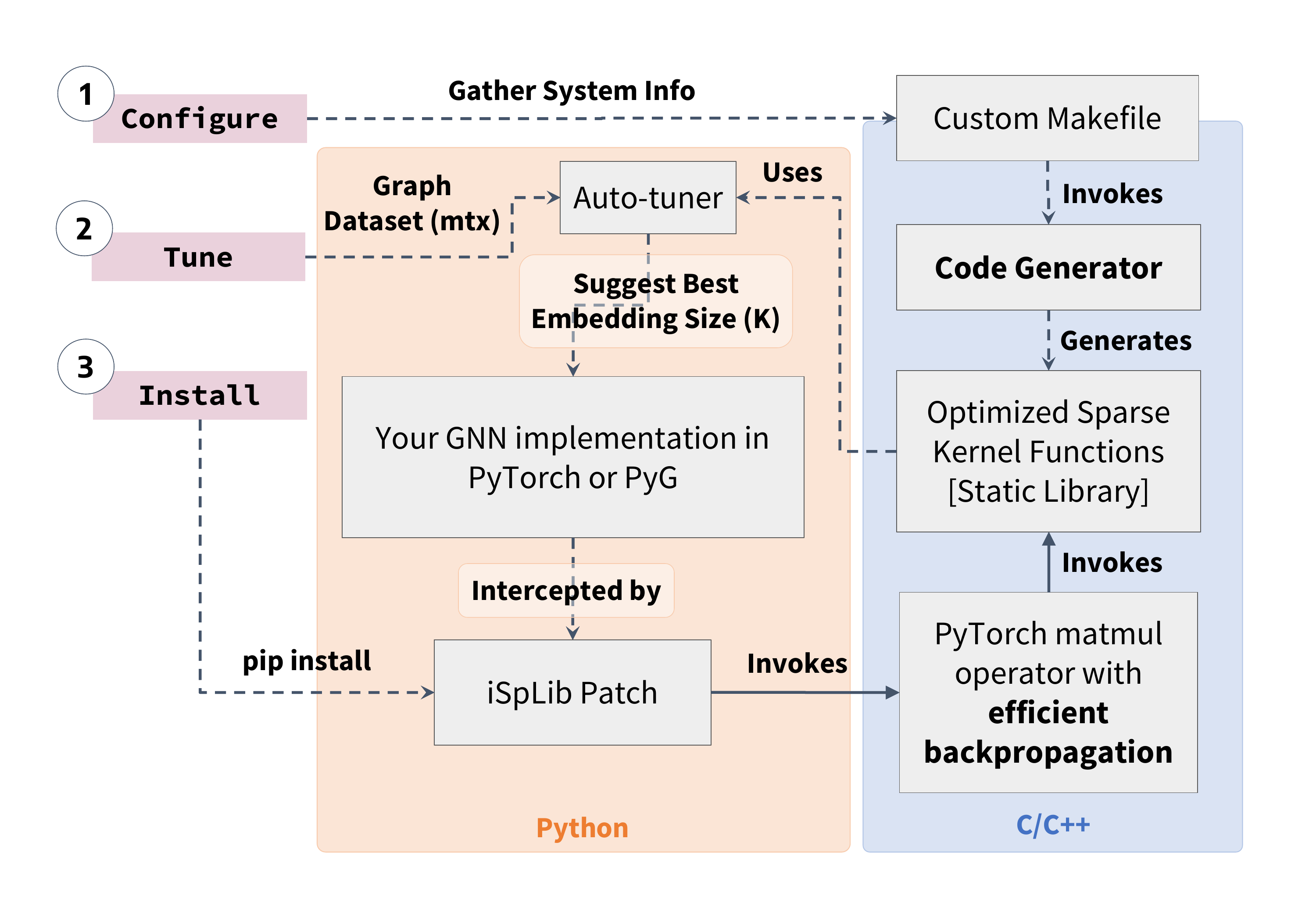} 
      \caption{Overview of the iSpLib Library}
      \label{fig:overview}
    
\end{figure}

The codebase of iSpLib consists of Python, C++, and C code (see Figure \ref{fig:overview}). The highly efficient kernels are generated in pure C. A C++ PyTorch wrapper connects the forward and backward propagation with the generated sparse kernels and provides the abstraction between PyTorch Tensors and the C++ array. A Python interface of iSpLib provides a ready-to-use matmul function for performing sparse-dense matrix multiplication. iSpLib also includes a PyTorch Geometric plug-in that allows users to use our matrix multiplication in popular GNN implementations when the dataset format is compatible.

\subsection{Auto-tuning Mechanism}
iSpLib has an auto-tuning mechanism that suggests the optimal embedding size for a given user environment. iSpLib probes the hardware to determine SIMD vector length and generates kernels for various multiples of these vector lengths (VLEN). 
When the embedding dimension is not a multiple of VLEN, we use a trusted kernel that is still efficient with balanced multithreading, but it does not use loop unrolling. The auto-tuning feature allows users to tune the library against a given dataset by generating a comparison chart for speedup on the generated kernels over the trusted kernels for a sequence of embedding sizes (K). Typically the tuning graph is a bell-shaped curve where the peak corresponds to the ideal embedding size (K) since it is where the generated kernel achieved the highest speedup compared to the trusted kernel.

\subsection{Efficient Backpropagation}
Another major source of iSpLib's speedup comes from the backpropagation. iSpLib's intelligent matrix-multiplication kernel is designed to identify common expressions required during the training epochs and cache them locally. This caching mechanism greatly reduces the time spent in backpropagation especially when the input graph is large or the number of epochs is high.

\subsection{Semiring Support}
The sparse-dense matrix multiplication of iSpLib also supports various semirings containing user-defined operations. This is particularly useful for training GraphSAGE that involves aggregation methods other than sum. iSpLib's matmul operator also supports min, max, and mean as reduction operations other than the standard sum operation. Currently, only the sum reduction operation has the generated kernel support in our library.

\subsection{Matmul Interface and Dependency}
iSpLib provides a PyTorch-based interface for sparse-dense matrix multiplication. The sparse parameter of the matmul function receives a SparseTensor\footnote{A sparse matrix data structure provided by pytorch\_sparse Python library} in CSR (compressed sparse row) format, the dense matrix as a typical 2-d PyTorch tensor, and optionally a reduction operation string having either `sum', `min', `max', or `mean' as input. Pytorch\_sparse library provides a transform function to convert existing PyTorch-based datasets into SparseTensor format. After dataset conversion, our matmul function can be used to develop any GNN that requires sparse-dense matrix multiplication.

\subsection{PyTorch Geometric Integration}

Finally, iSpLib provides a PyG `patch' and `unpatch' function that allows users to seamlessly integrate our auto-tuned matmul function into existing PyG implementation of GNNs that involve sparse-dense matrix multiplication. This can be done by importing the iSplib library at the top of the PyG implementation code and invoking the patch function. The users can also disengage the iSpLib interception at any point of code by invoking the `unpatch` method. iSpLib also provides a decorator for patching a single function in the PyG implementation.


\begin{table}
  \caption{Datasets}
  \label{dataset_table}
  \begin{tabularx}{0.48 \textwidth }{p{2.4cm}p{.8cm}p{1.2cm}p{1.1cm}p{1cm}p{5cm}}
\toprule
Graph Dataset        & Feature Length & Prediction class & Node Count & Edge Count  \\
\midrule

Reddit         & 602            & 41               & 232,965    & 11,606,919  \\
Reddit2        & 602            & 41               & 232,965    & 23,213,838  \\
OGBN-mag       & 128            & 349              & 736,389    & 135,680,469 \\
Amazon Products & 200            & 107              & 1,569,960  & 264,339,468 \\
OGBN-Product   & 100            & 47               & 2,449,029  & 61,859,140  \\
OGBN-Protein   & 8              & 112              & 154,154    & 159,462   \\ 
\bottomrule
\end{tabularx}
  
\end{table}

\section{Experimental Setting}
We measure the training time of various implementations of several two-layer GNNs on node classification tasks and compared against iSpLib. For comparison, we test model variants from PyTorch 2.1.0 (sparse), PyTorch < 2.0 (sparse), PyTorch 2 non-sparse (message passing) model, and PyTorch 2 torch.compile method against iSpLib. We select Graph Convolution Network (GCN), GraphSAGE (sum and mean), and Graph Inference Network (GIN) as they are widely used for benchmarking. 

We select six large graph datasets and performed a one-dim node prediction task for 30-100 epochs while measuring the average training time. We conduct all experiments in two high-configuration CPUs: (a) an Intel Skylake CPU with 48 cores and 256GB memory, and (b) an AMD EPYC 7763 64-core Processor with 527GB memory. The datasets used in our experiment are presented in Table \ref{dataset_table}.

\section{Results}
\begin{figure}[th]
    \begin{subfigure}{.33\textwidth}
      \includegraphics[width=1.1\linewidth]{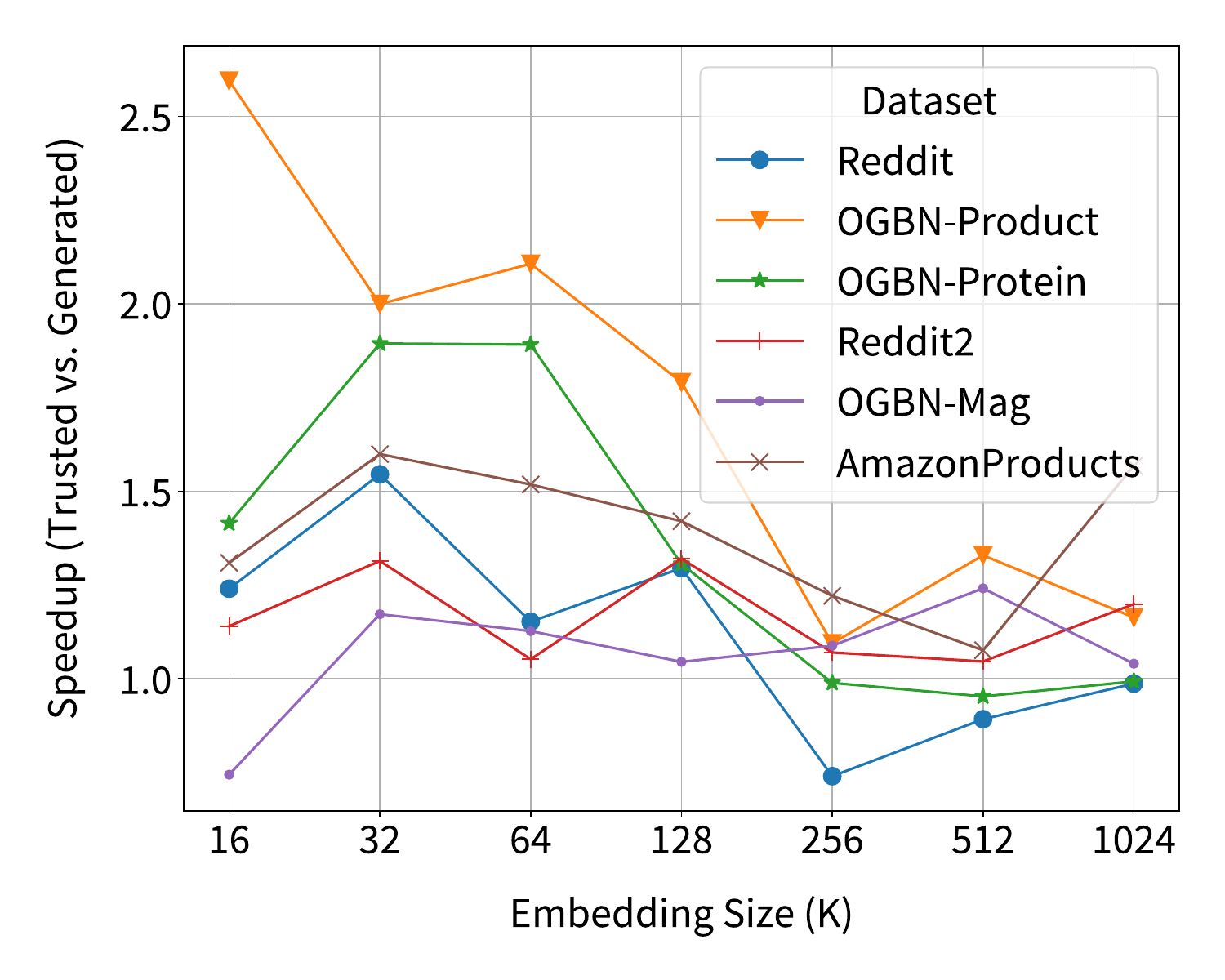}
      \caption{Intel}
      \label{fig:tuning-intel}
    \end{subfigure}
    \begin{subfigure}{0.33\textwidth}
      \includegraphics[width=1.1\linewidth]{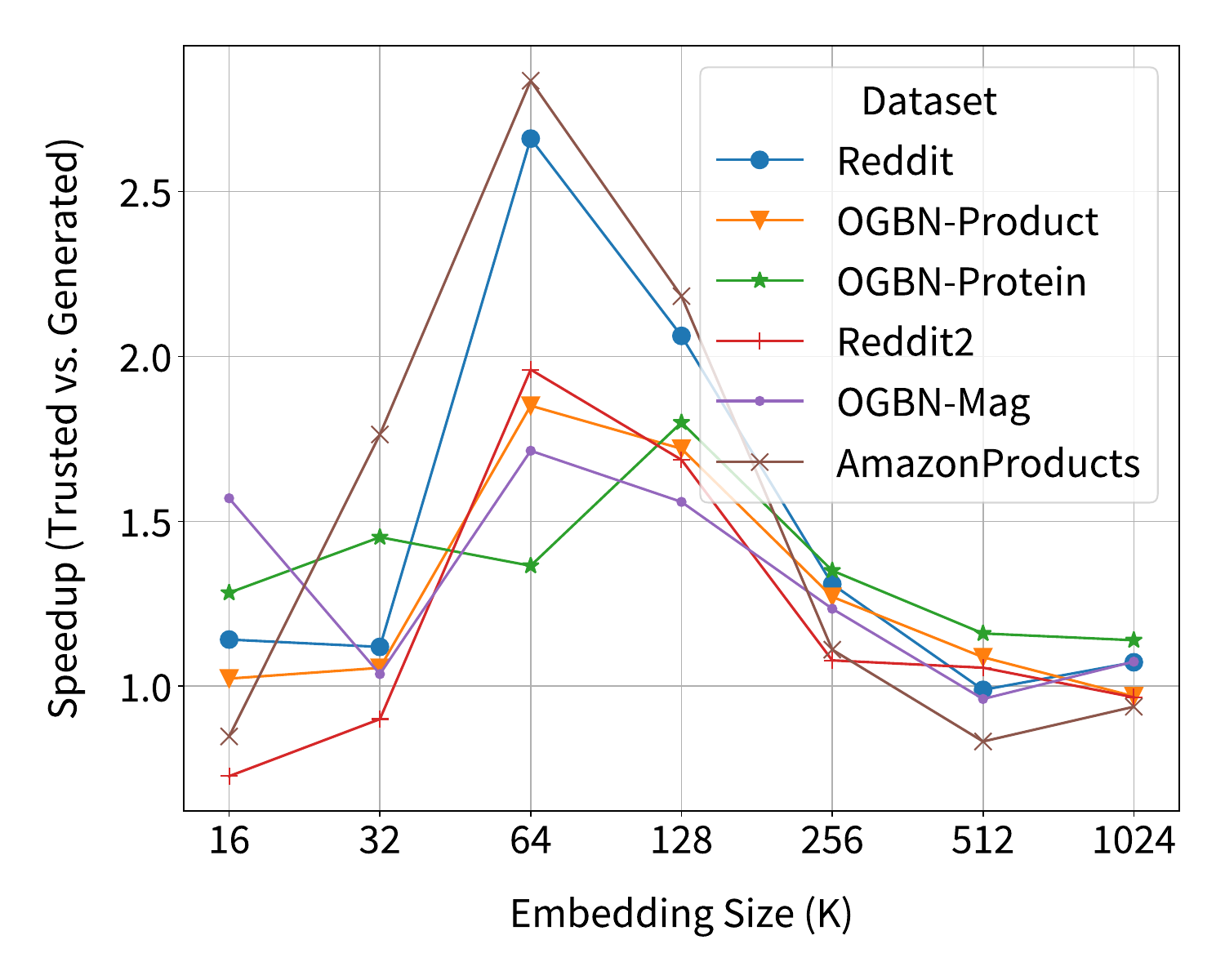}
      \caption{AMD}
      \label{fig:tuning-amd}
    \end{subfigure}
    \caption{Tuning Graph for various CPUs}
    \label{fig:tuning-graphs}
\end{figure}

{\bf Auto-tuning results.}
We generate the tuning graph for both Intel and AMD CPUs and show the result in Figure \ref{fig:tuning-graphs} for embedding sizes 16, 32, 64, 128, 256, 512, and 1024 for all six datasets. The figure is used to identify the most efficient embedding sizes for the corresponding CPUs (32 for Intel and 64 for AMD). We use these values to run the GNNs for both CPUs.

\begin{figure*}[th]
\begin{subfigure}{1.03\textwidth}
\includegraphics[width=1\linewidth]{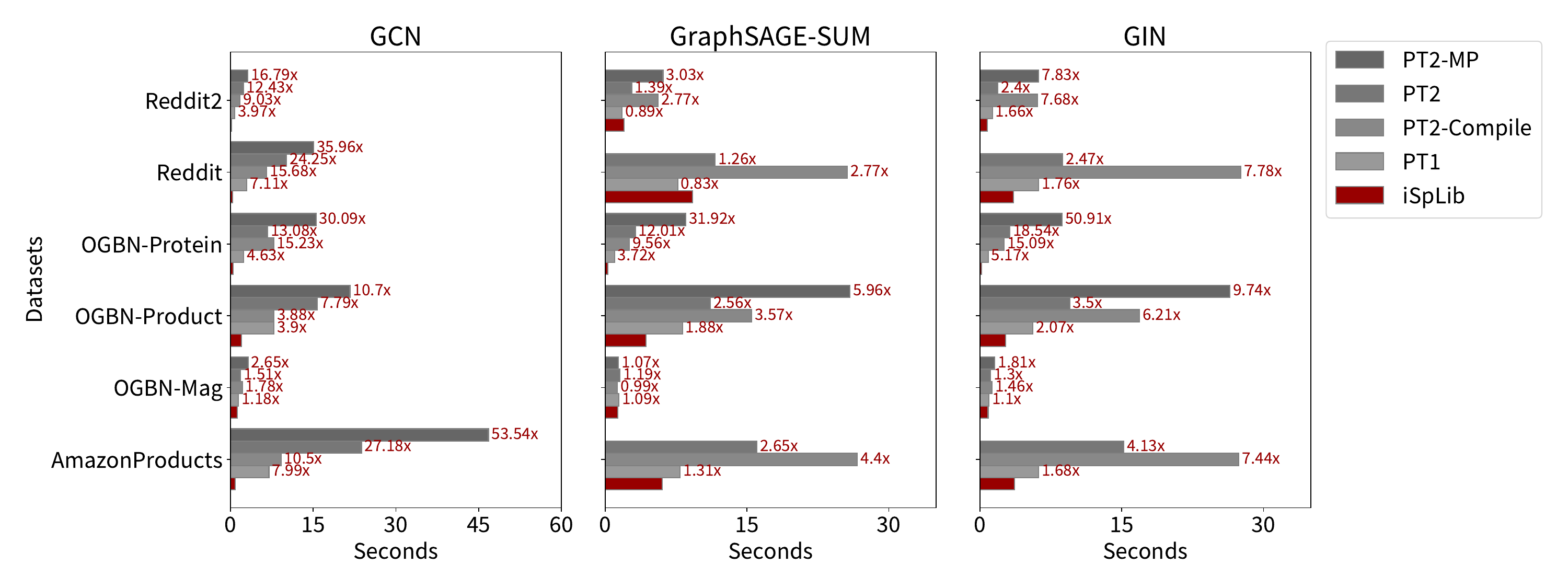}

      \label{fig:intel-speedup}
    
    \caption{Intel}
    \end{subfigure}

    \begin{subfigure}{1.03\textwidth}
\includegraphics[width=1.025\linewidth]{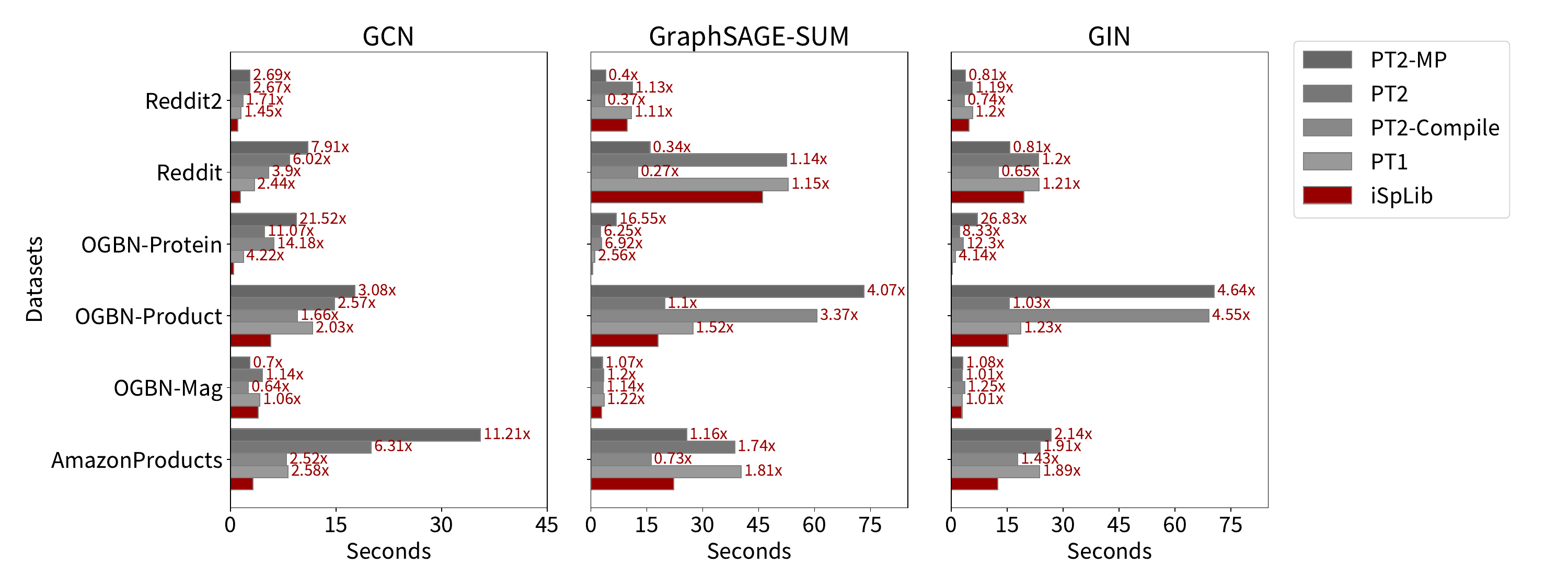}

      \label{fig:amd-speedup}
    
    \caption{AMD}
    \end{subfigure}
\caption{Average per-epoch training time and speedup for iSpLib w.r.t. other frameworks [PT2: PyTorch 2.1, PT1: PyTorch < 2, PT2-Compile: PyTorch 2 torch.compile, PT2-MP: PyTorch 2 Message Passing paradigm]}
\label{fig:speedups}
\end{figure*}

{\bf GNN training performance.}
We train three GNN models using iSpLib and four other settings and show the average per-epoch training time in Figure \ref{fig:speedups}. 
Due to the limitation of space, we omit the results for GraphSAGE-MEAN.
Since iSpLib is a drop-in replacement of PyTorch SpMM operation, it does not alter the results found in PyTorch. Thus the training and testing accuracy remains the same for all GNNs.
Figure \ref{fig:speedups} shows that iSpLib can accelerate GNN training significantly when compared with other frameworks. 
However, the speedup varies across GNN models and datasets. 

{\bf Performance across GNN models.}
Figure \ref{fig:speedups} shows that iSpLib achieves better speedup for GCN compared to GraphSAGE and GIN. This is because GCN typically performs a linear projection on the feature matrix before running the convolution. This step projects high-dimensional features into low-dimension space for which tuned kernels perform better (see Figure \ref{fig:tuning-graphs}).
GraphSAGE and GIN do not have the initial projection of the feature matrix and perform SpMM with original features. This makes iSpLib relatively less effective for GraphSAGE and GIN. 
However, for datasets that originally had a lower feature count in the feature matrix such as OGBN-Protein (feature size: 8), we observe GCN-like speedup in GraphSAGE and GIN for both Intel and AMD CPUs.

{\bf Comparison with other GNN frameworks.}
Additionally, we compare CogDL's \cite{cen2021cogdl} equivalent GCN implementation w.r.t. iSpLib and observe up to 43x speedup for various datasets. We also compare iSpLib with PyTorch 2.1 on the vanilla implementation of GCN and observe up to 93x speedup for the Reddit dataset on Intel CPU.

\section{Discussion}
We observe that the embedding size suggested by the autotuner is low, i.e., we have a better chance of seeing an improved performance from generated kernels for smaller embedding sizes. This is due to the fact that iSpLib's generated kernels perform register blocking to reduce cache misses. For larger embedding sizes, it has to allocate more values to the register, causing register spilling and increased cache misses.



We see less overall speedup for OGB-Mag since it is a smaller graph compared to others. Caching a smaller graph has less impact on the speedup in backpropagation, thus caching the expressions does not improve the training time significantly. But still, we observe better performance in iSpLib compared to the other frameworks for most scenarios due to the usage of efficient kernels.

\section{Conclusion}
We developed a user-friendly sparse matrix library called iSpLib that works as a drop-in replacement for existing PyTorch and PyG equivalent operations. iSpLib provides auto-tuned and customized kernels for target user environments.  The library also supports backpropagation that caches repetitive data during the training phase. We observe up to $27\times$ speedup w.r.t. equivalent PyTorch 2 implementation for training larger graphs on popular GNNs.

\begin{acks}
This research is supported by the NSF OAC-2112606 and OAC-2339607 grants and DOE DE-SC0022098 and DE-SC0023349 awards. 
We extend our appreciation to the PyTorch\_Sparse community for open-sourcing their project. Our project has significantly benefited from their coding style and implementation.

\end{acks}

\bibliographystyle{ACM-Reference-Format}
\bibliography{sample-base}

\appendix

\end{document}